# Building machines that adapt and compute like brains

Nikolaus Kriegeskorte and Robert M. Mok

*Medical Research Council Cognition and Brain Sciences Unit*

*University of Cambridge, 15 Chaucer Road, CB2 7EF Cambridge, United Kingdom*

n.kriegeskorte@columbia.edu, robmmok@gmail.com

*Abstract*: **Building machines that learn and think like humans is essential not only for cognitive science, but also for computational neuroscience, whose ultimate goal is to understand how cognition is implemented in biological brains. A new cognitive computational neuroscience should build cognitive-level and neural-level models, understand their relationships, and test both types of models with both brain and behavioral data.**

Lake et al.'s timely commentary puts the recent exciting advances with neural network models in perspective, and usefully highlights the aspects of human learning and thinking that these models do not yet capture. Deep convolutional neural networks have conquered pattern recognition. They can rapidly recognize objects as humans can, and their internal representations are remarkably similar to those of the human ventral stream (Eickenberg, Gramfort, Varoquaux, & Thirion, 2016; Güçlü & van Gerven, 2015; Khaligh-Razavi & Kriegeskorte, 2014; Yamins et al., 2014). However, even at a glance, we understand visual scenes much more deeply than current models. We bring complex knowledge and dynamic models of the world to bear on the sensory data. This enables us to infer past causes and future implications, with a focus on what matters to our behavioral success. How can we understand these processes mechanistically?

The top-down approach of cognitive science is one required ingredient. Human behavioral researchers have an important role in defining the key challenges for model engineering by introducing tasks where humans still outperform the best models. These tasks serve as benchmarks, enabling model builders to measure progress and compare competing approaches. Cognitive science introduced task-performing computational models of cognition. Task-performing models are also essential for neuroscience, whose theories cannot deliver explicit accounts of intelligence without them (Eliasmith & Trujillo, 2014). The current constructive competition between modeling at the cognitive and neural levels, is inspiring and refreshing. We need both levels of description to understand, and to be able to invent, intelligent machines and computational theories of human intelligence.

Pattern recognition was a natural first step toward understanding human intelligence. This essential component mechanism has been conquered by taking inspiration from the brain. Machines could not do core object recognition (DiCarlo, Zoccolan, & Rust, 2012) until a few years ago (Krizhevsky, Sutskever, & Hinton, 2012). Brain-inspired neural networks gave us machines that can recognize objects robustly under natural viewing conditions. As we move toward higher cognitive functions, we might expect that it will continue to prove fruitful to think about cognition in the context of its implementation in the brain. To understand how humans learn and think, we need to understand how brains adapt and compute.

A neural network model may require more time to train than humans. This reflects the fact that current models learn from scratch. Cognitive models, like Bayesian program learning (Lake, Salakhutdinov, & Tenenbaum, 2015), rely more strongly on built-in knowledge. Their inferences require realistically small amounts of data, but unrealistically large amounts of computation, and, as a result, their high-level feats of cognition don't always scale to complex real-world challenges. To explain human cognition, we must care about efficient implementation and scaleability, in addition to the goals of computation. Studying the brain can help us understand the representations and dynamics that support the efficient implementation of cognition (e.g. Aitchison & Lengyel, 2016).

The brain seamlessly merges bottom-up discriminative, and top-down generative processes into a rapidly converging process of inference that combines the advantages of both: the rapidity of discriminative inference and the flexibility and precision of generative inference (Yildirim, Kulkarni, Freiwald, & Tenenbaum, 2015). The brain's inference process appears to involve recurrent cycles of message passing at multiple scales, from local interactions within an area to long-range interactions between higher- and lower-level representations.

As long as major components of human intelligence are out of reach of machines, we are obviously far from understanding the human brain and cognition. As more and more component tasks are conquered by machines, the question of whether they do it "like humans" will come to the fore. How should we define "human-like" learning and thinking? In cognitive science, the empirical support for models comes from behavioral data. A model must not only reach human levels of task performance, but also predict detailed patterns of behavioral responses (e.g. errors and reaction times on particular instances of a task). However, humans are biological organisms, and so "human-like" cognition should also involve the same brain representations and algorithms that the human brain employs. A good model should somehow match the brain's dynamics of information processing.

Measuring the similarity of processing dynamics between a model and a brain, has to rely on summary statistics of the activity, and may be equally possible for neural and cognitive models. For neural network models, a direct comparison may seem more tractable. We might map the units of the model onto neurons in the brain. However, even two biological brains of the same species will have different number neurons, and any given neuron may be idiosyncratically specialized, and may not have an exact match in the other brain. For either a neural or a cognitive model, we may find ways to compare the internal model representations to representations in brains (e.g.

Kriegeskorte & Diedrichsen, 2016; Kriegeskorte, Mur, & Bandettini, 2008). For example, one could test whether the visual representation of characters in high-level visual regions reflects the similarity predicted by the generative model of character perception proposed by Lake et al. (2015).

The current advances in AI reinvigorate the interaction between cognitive science and computational neuroscience. We hope that the two can come together and combine their empirical and theoretical constraints, testing cognitive and neural models with brain and behavioral data. An integrated cognitive computational neuroscience might have a shot at the task that seemed impossible a few years ago: understanding how the brain works.